\documentclass[runningheads]{llncs}
\usepackage[T1]{fontenc}
\usepackage{graphicx}
\usepackage{booktabs}
\usepackage{amsmath, amssymb}
\usepackage{pdflscape}
\usepackage{rotating}

\usepackage{comment}

\usepackage{color}

\begin{document}
\title{Learning Cardiac Features: ECG Biometrics Across Time and Exercise}
\author{
Luca Thiebaud\inst{1} \and
Paul Chauchat\inst{1} \and
Mustapha Ouladsine \inst{1} \and 
Stéphane Delliaux \inst{2, 3} 
}
\authorrunning{L. Thiebaud et al.}
\institute{
Aix-Marseille Univ, CNRS, LIS, Marseille, France \and
Aix-Marseille Univ, Inserm-INRAE, C2VN, Marseille, France \and
University Hospitals of Marseille, Marseille, France
}

\maketitle

\begin{abstract}

Electrocardiograms (ECGs) carry subject-specific patterns enabling reliable individual discrimination, forming the basis of ECG biometrics. Beyond authentication, this paradigm holds significant potential to secure sensitive cardiac data and to serve as a pretext task in self-supervised learning. Yet, most studies remain confined to single-session, resting data, leaving robustness to temporal and physiological variations largely untested. We address this gap by evaluating ECG biometrics under realistic conditions involving exercise-induced stress and cross-session variability. A Siamese ResNet with late multi-lead fusion strategy is trained on a large ECG dataset extracted from cardiopulmonary exercise tests and evaluated with a exercise- and time-aware protocol, as well as on public benchmarks. This first extensive assessment of ECG biometrics under combined physiological and temporal variability achieves an intra-session rest-to-peak EER of 1.7\% and state-of-the-art 3.9\% on the CYBHi dataset. 
Findings support the presence of an intrinsic cardiac signature resilient to physiological and temporal drift.  
\end{abstract}

\keywords{ECG biometrics \and Siamese networks \and Deep learning \and Exercise stress \and CPET}

\section{Introduction}

The electrocardiogram (ECG) is a fundamental tool for assessing cardiac electrical activity \cite{Kligfield2007}. While its analysis has traditionally depended on expert interpretation, machine learning—and particularly deep learning—has greatly advanced automatic ECG-based diagnosis and classification \cite{Minchole2019,Liu2021}, notably through ResNet architectures \cite{Chen2025,pmlr-nonaka21a}. To overcome the reliance on labeled data, recent work has turned to self-supervised learning (SSL) based approaches, enabling pre-training large architectures without the need for large amounts of annotated data \cite{Zhang2024}. In this context, ECG biometrics has emerged as a ECG SSL pretext task \cite{Diamant2022,Chen2025}, enhancing downstream clinical tasks such as pathology classification. 

Beyond model training, ECG-based biometrics also stand as a secure and physiology-rooted alternative to conventional face or fingerprint recognition methods \cite{Melzi2023}. This raises critical privacy implications at the same time, as even anonymized ECG traces can reveal identifying information across large datasets \cite{Macierzanka2025}. Therefore, ECG biometrics appears both as a compelling pretext task for downstream medical purposes, and as a crucial privacy-related approach in itself which should be thoroughly analyzed.

Deep learning methods have become the state-of-the-art in ECG-based biometrics, outperforming classical fiducial and handcrafted feature approaches \cite{Melzi2023,Meltzer2025}. However, methodological heterogeneity—differences in tasks (identification vs.\ authentication), dataset scale, acquisition, or evaluation—still impedes objective benchmarking \cite{Melzi2023}. Identification systems (one-to-many, 1:N) typically require retraining when new users are added, limiting scalability, while authentication protocols (one-to-one, 1:1) enable verification without retraining and better reflect real-world use. Yet, most existing works still rely on identification settings with small populations, often under a few hundred subjects \cite{Meltzer2025}.

Another persistent weakness lies in temporal variability. Many studies rely on intra-session evaluation (training and testing within the same recording period), reporting near-perfect accuracies that reflect session consistency more than biometric permanence \cite{Melzi2023}. In contrast, the few cross-session analyses (i.e., training and testing on recordings acquired days to years apart) \cite{Ibtehaz2022,Chee2022,Melzi2023,Islam2022} reveal substantial performance degradation over time, questioning both the permanence of ECG as a biometric trait and the quality of the learned representations.

Finally, the impact of physiological ECG changes on identifiability remains largely unexplored. Most datasets are recorded at rest, and the few studies examining exercise or emotional stress \cite{Zhou2021,Sung2017,Komeili2016,Cui2021} report substantial drops in recognition accuracy, highlighting limited robustness to physiological variability.

These limitations highlight the need for large, diverse datasets covering temporal and physiological variability, and for learning strategies able to extract invariant ECG representations resilient to context-dependent changes.\\

In this study, we propose a deep learning–based ECG biometric authentication method and assess its performance under cross-session and exercise-induced variability using a large-scale dataset. Our main contributions are as follows:
\begin{itemize}
    \item A Siamese ResNet with late multi-lead fusion that improves robustness to cross-session variability compared to conventional early fusion schemes
    \item The first extensive evaluation of ECG biometric stability under exercise stress, showing consistent performance from rest to exertion
    \item State-of-the-art authentication results, including cross-dataset evaluations, surpassing existing methods
\end{itemize}
The paper is organized as follows: Section \ref{sec:related} reviews related work; Section \ref{sec:CPET_Dataset} describes the CPET dataset; Section \ref{sec:methods} details the methodology; Section \ref{sec:experiments} presents experiments and results; and Section \ref{sec:conclusion} concludes the study.

\section{Related Works}
\label{sec:related}

\subsection{Evolution of ECG Biometric Methodologies: From Fiducial to Deep Models}

Early ECG biometrics employed fiducial approaches, extracting waveform landmarks (P-QRS-T peaks, RR intervals) \cite{Fratini2015} and non-fiducial methods using autocorrelation or wavelet transforms \cite{Meltzer2025}. Both suffer from noise sensitivity, imprecise peak detection, and limited robustness to cross-session variability and physiological changes. Recent years have seen deep learning methods dominate ECG biometrics, leveraging raw signals to overcome fiducial limitations \cite{Melzi2023}.

Identification-driven works use standard CNN architectures. The ECG can be turned into an image to allow the use of standard image processing tools: Depth-wise separable convolutions showed good performance in intra-session arrhythmic beat identification \cite{AlJibreen2024}, and an ensemble of fine-tuned pretrained ResNets and DenseNets showed very good performance in a cross-session setting \cite{Srivastva2021}. While a transformer-based approach have produced good authentication performances \cite{Chee2022}, almost all methods rely on Siamese networks \cite{Meltzer2025}. The backbone can be pretrained on identification \cite{Ibtehaz2022} or autoencoding tasks \cite{Melzi2023}.

\subsection{Methodological Limitations}

\subsubsection{Cross- vs. Intra-session}
Several recent reviews show that most ECG biometric studies are still conducted in intra-session settings and report excellent performance. However, such protocols may conflate subject-specific cardiac traits with session-dependent acquisition factors, including electrode placement, posture, or psychophysiological state. This raises the question of whether intra-session results fully reflect stable biometric properties or instead partly capture session-specific effects \cite{Meltzer2025,Melzi2023}.

In contrast, more recent works \cite{Melzi2023,Ibtehaz2022,Chee2022,Islam2022} explicitly address this limitation by enforcing strict cross-session evaluation splits and consistently reporting the associated performance degradation. Moreover, \cite{Melzi2023} introduced a benchmarking framework encompassing both mono- and multi-session settings across multiple databases, advancing the standardization of ECG biometric evaluation.

\subsubsection{Identification vs. Authentication}
Finally, many works rely on identification rather than authentication protocols \cite{Srivastva2021,AlJibreen2024}. Identification (1:N) aims at determining the identity of a subject among a fixed set of enrolled users, whereas authentication (1:1) verifies a claimed identity by comparing two biometric samples. Identification thus generally requires retraining when new individuals are added, limiting scalability, while authentication approaches allow verification without retraining. 

Moreover,  authentication enables more realistic evaluation protocols such as patient-level splits, where testing is performed on individuals not seen during training. This clearly transpires in cross-session studies, where identification results easily reach 99\% or higher. However, in this setting, the associated networks are trained on a first session of each patient, therefore patients whose second session lies in the test set were already seen in the training phase, inducing data leakage \cite{Srivastva2021}.

Although more robust, cross-session authentication is still rarely explored in the literature. Notably, \cite{Srivastva2021} reported that among 21 studies (yielding 32 results on authentication and identification tasks across various ECG databases), only five results correspond to cross-session authentication protocols. %This is why we focus on this specific setting in this work.

\subsection{Dataset Limitations}

\subsubsection{Population Size}
Most ECG-biometrics studies rely on limited sample sizes. As emphasized in \cite{Meltzer2025}, “existing works on ECG for user authentication do not consider a population size close to a real application.” Typically, datasets include only a few dozen participants and seldom exceed a few hundred individuals.

\subsubsection{Impact of Physiological Variability on Biometric Performance}
\label{sec:existing_works_stress}
Most studies on ECG biometrics rely on resting recordings, the standard condition in available databases. Only a few works have investigated performance under varying physiological conditions \cite{Zhou2021,Sung2017,Komeili2016,Cui2021}. These studies are generally based on small datasets (20–70 subjects), are limited to identification tasks, and consistently report marked degradation when comparing rest and non-resting conditions.

In \cite{Sung2017}, recognition accuracy dropped to approximately 60\% when enrollment was performed at rest and verification immediately after exertion, before recovering above 90\% within one minute and 96\% within five minutes, highlighting the impact of post-exercise physiological recovery. Similarly, \cite{Cui2021} reported a decrease from 99.7\% (rest–rest) to 88\% (exercise–exercise), and only 17.5\% under cross-condition evaluation (rest–exercise).

Overall, the current state of the art highlights the lack of robustness of ECG biometrics under physiological stress, compounded by methodological limitations such as small dataset sizes and the predominant use of identification rather than authentication protocols.

\section{CPET Dataset}
\label{sec:CPET_Dataset}

\subsection{Study Population}
\label{sec:cohort}
The data were collected from the Pulmonary Function Testing Laboratory of the Hôpital Nord, Assistance Publique – Hôpitaux de Marseille, between 2010 and 2024, with approval from the Clinical Ethics Committee. All participants underwent symptom‑limited incremental exercise tests on an ergocycle, performed as part of routine cardiopulmonary evaluation. A total of 1651 adult patients were included in the study (mean age = 58.5 $\pm$ 14.8 years; 59.4\% males). Among them, 1523 participants underwent a single CPET (mean age = 58.9 $\pm$ 14.7 years; 58.9\% males), while 128 participants completed multiple sessions during the study period (mean age = 53.8 $\pm$ 15.2 years; 65.6\% males).

\subsection{CPET Protocol and Data Acquisition}

\subsubsection{Exercise Protocol}

\begin{figure}[ht]
    \centering
    \includegraphics[width=1\linewidth]{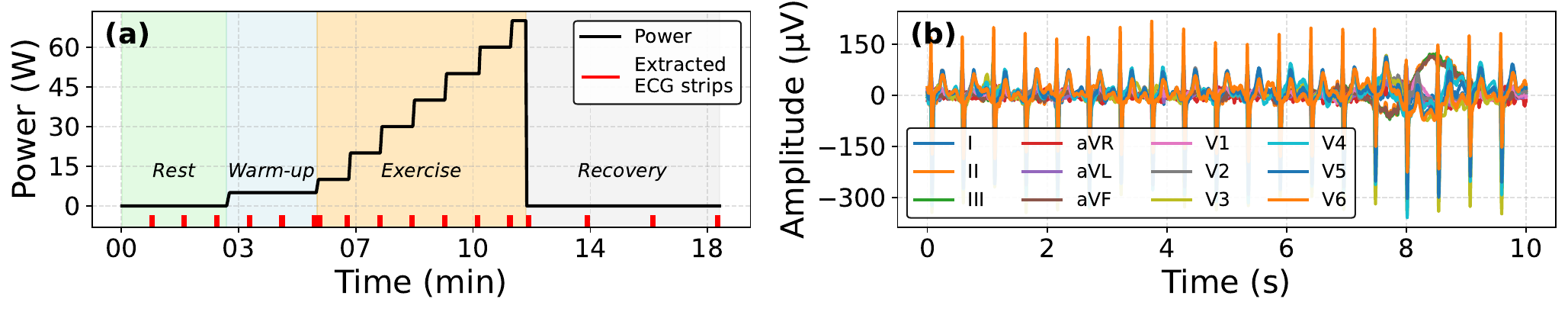}
    \caption{CPET protocol and ECG extraction. (a) Incremental workload with 10s ECG strips (red) sampled during the test. (b) Example 12-lead ECG strip during the incremental phase.}
    \label{fig:cpet_strip}
\end{figure}

CPET is performed on a cycle ergometer following a standard incremental protocol \cite{Balady2010}, comprising four phases (Fig.~\ref{fig:cpet_strip}(a)): rest (baseline recording, no cycling), warm-up (low constant workload), incremental exercise (progressive ramp to exhaustion or clinical stop), and recovery (post-exercise monitoring until return toward baseline).

\subsubsection{ECG Acquisition During CPET}
For medical and patient security purposes, ECG is continuously monitored during CPET using CardioSoft v7.0 system (GE Healthcare).
For each session, continuous 10-second 12-lead ECG strips are extracted throughout the test to provide balanced coverage of the CPET phases (see Fig.~\ref{fig:cpet_strip}(a)). The signal sampling rate is 500 Hz. Due to the nature of the exercise test, the signals can be noisy with motion artifacts (Fig.~\ref{fig:cpet_strip}(b)).

\subsection{ECG Preprocessing and Beat Extraction}
\label{sec:preproc}

\subsubsection{Signal Filtering and Segmentation}
All ECG signals are processed using tools from the NeuroKit2 Python library \cite{Makowski2021}. Noise removal is performed using a 5th-order Butterworth high-pass filter at 0.5 Hz combined with 50 Hz powerline filtering to correct baseline drift and reduce electrical interference.
R-peaks are then detected using the NeuroKit2 detection algorithm. Individual heartbeats are segmented around each detected R-peak using a fixed-length window of 0.8 s (0.32 s before and 0.48 s after the peak), as in \cite{Melzi2023}. This segmentation strategy has been shown to provide more stable morphological alignment than random cropping \cite{Li2020}. All extracted beats are finally standardized using z-score normalization prior to model training.

\subsubsection{Quality-based Beat Exclusion}
Due to the nature of the exercise test, signals can be too noisy or exhibit motion artifacts. We assessed beat quality using the template-matching quality index \cite{Makowski2021}, discarding beats below 85\%. This threshold, selected based on visual inspection and kept fixed without further optimization, allowed retaining more than 70\% of the beats.

\subsection{Comparison with Existing ECG Databases}

Public ECG datasets suitable for cross-session biometric evaluation remain scarce and are mostly recorded at rest, notably PTB~\cite{Bousseljot2004}, CYBHi~\cite{Silva2014} and Heartprint~\cite{Islam2022}, which we use for comparison. PTB provides 12-lead recordings with a short median inter-session interval (MISI) of 4 days, whereas CYBHi and Heartprint use single-lead finger acquisitions with MISIs of about 3 months and 4 years, respectively. For Heartprint, only the S1--S3L configuration was considered to maximize the inter-session interval. ECG-ID~\cite{Lugovaya2011} also enables cross-session analysis but was excluded due to its shorter MISI, while MIT-BIH~\cite{Moody1992} and \cite{DeGiovanni2021} contain only one session per subject.
As shown in Table~\ref{tab:dataset_characteristics}, the CPET dataset differs markedly from public ECG databases by combining a larger cohort, repeated exercise recordings, and a MISI of 346 days, whereas most public datasets are limited to rest. For all datasets, only the first two recordings were retained when more than two were available, following \cite{Melzi2023}.

\begin{table}[t]
\centering
\caption{ECG datasets used in this study. $^*$ MISI: Median Inter-Session Interval. $^\dagger$ Finger-to-finger. Unlike public datasets, mostly recorded at rest, CPET includes varying physiological conditions.}

\setlength{\tabcolsep}{0.0065\linewidth}
\begin{tabular}{lccccccc}
\toprule
 & \textbf{CPET} & PTB & CYBHi & Heartprint & ECG-ID & \cite{DeGiovanni2021} & MIT-BIH \\
\midrule
Total indiv. & 1651 & 290 & 63 & 78 & 90 & 22& 423\\
Indiv. with $\geq$2 sess. & 128 & 113 & 63 & 78 & 90 & 0 & 0 \\
MISI\textsuperscript{*} (days) & 346 & 4 & 104 & 1572 & Same day & - & - \\
ECG leads & 12 & 12 & $1^\dagger$ & $1^\dagger$ & $1^\dagger$ & 12 & 12 \\
% Acquisition & Chest & Chest & Fingers & Chest & Chest & Chest \\
Recording condition & Varying & Rest & Rest & Rest & Rest & Exercise & Rest \\
\bottomrule
\end{tabular}
\label{tab:dataset_characteristics}
\end{table}

\subsection{Exercise-Induced ECG Morphological Variability}

\begin{figure}[ht]
    \centering
    \includegraphics[width=1\linewidth]{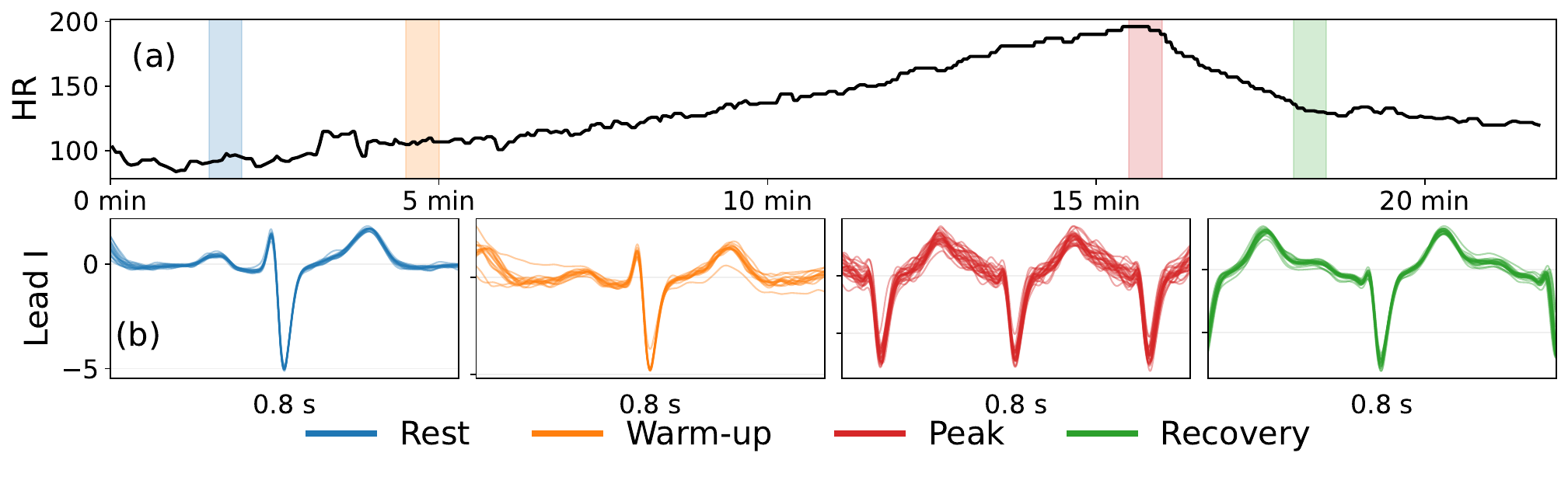}
    \caption{For one patient. (a) Heart rate (HR, bpm) over the test. (b) Lead-I ECG morphology across phases (0.8 s, Z-score normalized), showing exercise-induced changes.}
    \label{fig:HR_change}
\end{figure}

During incremental exercise, cardiac electrical activity evolves according to physiological adaptation, mainly driven by the autonomic regulation. The muscularly and neurally induced cardiovascular and hemodynamic changes associated with exercise primarily lead to heart rate increase. (Table~\ref{tab:hr_stats}) and signal deformation. Figure~\ref{fig:HR_change} illustrates intra-session variability for one patient: HR rises progressively (top), while ECG morphology across leads (bottom) shows P-wave amplification, R-wave attenuation with QRS-axis shift, and ST–T alterations partially recovering post-exercise \cite{Simoons1975}. With a fixed 0.8\,s window, cycle shortening at high intensity leads to multiple beats per segment, introducing additional intra-session variability that challenges biometric recognition.

\vspace{-0.02 \linewidth}
\begin{table}[h]
\centering
\caption{Mean heart rate (HR, bpm) across subjects during CPET phases.}
\setlength{\tabcolsep}{10pt}
\begin{tabular}{cccccc}
\toprule
 & Rest & Warmup & VT & Peak & Recovery \\
\midrule
HR ($\mu${ \scriptsize$\pm\sigma$}) & 85 { \scriptsize$\pm 15$} & 93 { \scriptsize$\pm 15$} & 111 { \scriptsize$\pm 20$} & 136 { \scriptsize$\pm 25$} & 108 { \scriptsize$\pm 19$} \\
\bottomrule
\end{tabular}
\label{tab:hr_stats}
\end{table}

\section{Methods}
\label{sec:methods}

\subsection{Siamese ResNet Architecture}
\label{sec:siamese_arch}

We adopt a Siamese architecture, given its state-of-the-art performance in authentication and its ability to reuse the backbone as a generic feature extractor for downstream medical tasks, a key motivation of this work. The backbone is a 1D ResNet-18 \cite{He2015}, a strong lightweight baseline for ECG analysis~\cite{Chen2025,pmlr-nonaka21a}.

\begin{figure}[htbp]
    \centering
    \includegraphics[width=1\linewidth]{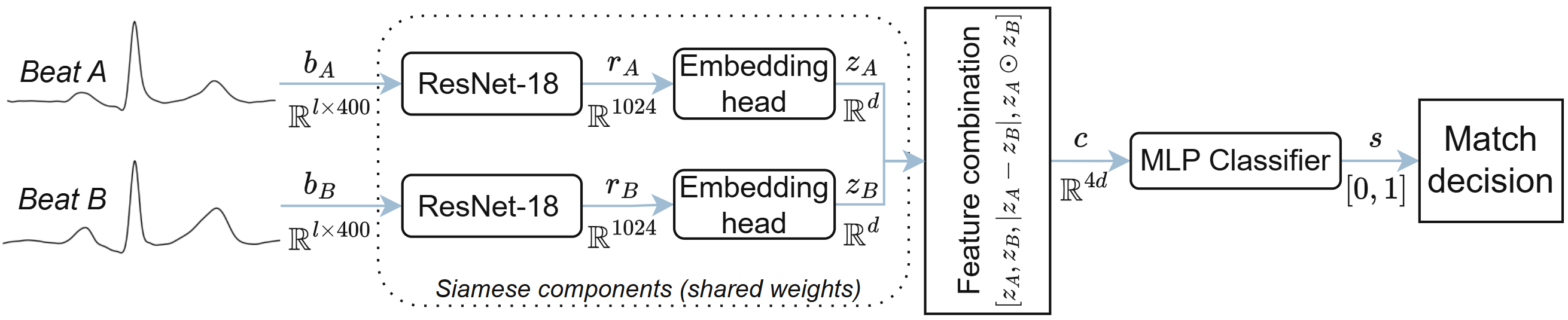}
    \caption{Overview of the proposed Siamese ResNet1D for ECG verification. Input beats have shape $(l,400)$, with $l$ leads and 400 samples ($0.8$\,s at 500\,Hz).}
    \label{fig:siamese}
\end{figure}

As shown in Fig.~\ref{fig:siamese}, two beats $b_A, b_B \in \mathbb{R}^{(l,400)}$ (with $l$ leads, of length $0.8\,\mathrm{s}$ at $500\,\mathrm{Hz}$) are processed independently by a shared-weight backbone. Feature maps are aggregated via adaptive concatenated pooling (max + average) into fixed-length vectors $r_A, r_B \in \mathbb{R}^{1024}$, then projected through a shared head (FC $1024\!\rightarrow\!d$, batch norm, dropout, ReLU) to embeddings $z_A, z_B \in \mathbb{R}^d$.

The embeddings are combined as $[z_A, z_B, |z_A - z_B|, z_A \odot z_B] \in \mathbb{R}^{4d}$, capturing both absolute and multiplicative interactions. A classifier (FC $4d\!\rightarrow\!256$, ReLU, batch norm, dropout, then FC $256\!\rightarrow\!1$) followed by a sigmoid outputs the score $s$ that both segments belong to the same individual.

\subsection{Single and Multi-lead Networks}
We explored three versions based on the architecture described in Section~\ref{sec:siamese_arch}. Two direct applications to single- and multi-lead inputs. For the multi-lead case, we propose a score fusion method, based on an ensemble of single-lead networks.

\subsubsection{Single-lead Architecture}
The single-lead network follows the Siamese architecture of Section~\ref{sec:siamese_arch} with $l=1$. The embedding dimension is $d=256$, and the ResNet-18 uses kernel size 3, selected via hyperparameter (HP) optimization.

\subsubsection{Multi-lead Early Fusion: ML-Early}
The multi-lead early fusion model uses the same architecture with input dimension $l$ equal to the number of leads. HP optimization yields $d=512$ and kernel size 7.

\subsubsection{Multi-lead Late Fusion: ML-Late}

ECG leads exhibit strong session-dependent correlations (e.g., electrode placement, recording conditions), so joint processing may induce overfitting, with the model capturing electrode configuration rather than subject-specific features. Following \cite{Aublin2022}, we propose a score-level fusion (ML-Late), based on several independent single-lead Siamese networks $s_i$, each trained solely on its respective lead $i$.
% , process the input beat pairs separately. The resulting match probabilities are then averaged to yield the final verification score (Fig.~\ref{fig:lateMl}).

Let $b_A, b_B \in \mathbb{R}^{(l,400)}$ be two multi-lead beats, and $b_{A,i}, b_{B,i} \in \mathbb{R}^{400}$ their $i$-th lead. Each single-lead Siamese expert $s_i$ produces a match score $s_i(b_{A,i}, b_{B,i}) \in [0,1].$
The final verification score is obtained by averaging across leads:
\[
s(b_A, b_B) = \frac{1}{l} \sum_{i=1}^{l} s_i(b_{A,i}, b_{B,i}).
\]

\subsubsection{Lead Selection Strategy for Multi-lead ECG}
\label{sec:Ml}

For standard 12-lead ECG, we retain eight leads (I, II, V1--V6) and omit III, aVR, aVL, and aVF, as they are linear combinations of I and II, following Einthoven's and Goldberger's relationships \cite{Ramirez2024}. Preliminary experiments show no performance gain from their inclusion, while increasing computational cost.

The multi-lead late fusion framework extends to other acquisition setups (e.g., finger-to-finger single-lead ECG) that do not match standard leads. We propose averaging scores from a subset of standard leads selected via geometry-aware criteria, based on the angular deviation between lead direction and the recording axis derived from standard electrode geometry \cite{Ramirez2024}.

\section{Experiments and Results}
\label{sec:experiments}

All experiments were conducted under deterministic conditions (fixed seeds), with models trained on the CPET database. Intra- and cross-session evaluations on CPET highlight the superiority of late fusion. Generalizability is then assessed on public datasets without fine-tuning, and compared to state-of-the-art methods. Finally, the impact of exercise-induced stress is analyzed.

\subsection{Experimental Protocol}
\label{sec:protocol}

\subsubsection{Data Splitting Strategy and Pair Construction}

The CPET dataset includes 1,523 single-session and 128 multi-session patients. To prevent data leakage, splitting was performed at the patient level, enabling both intra- and cross-session evaluations. Multi-session patients were evenly split (64/64), and single-session patients were split 85\%/15\%, yielding 82\% of patients for training, thereby limiting cross-session exposure during training while preserving it for testing.

Pairs were constructed at the beat level by pairing each heartbeat with another from the same subject (genuine; different session if available, otherwise different strip, possibly within the same phase of the CPET) and with a beat from another subject (impostor; random strip). The training set comprises 720,254 pairs balanced (1:1), used identically for 1-lead and 8-lead Siamese ResNet models, differing only in input dimensionality. The same procedure yields 146,472 single-session test pairs and 70,118 cross-session test pairs for testing (from 262 single-session and 64 multi-session test patients, respectively). For public datasets, strict benchmark compliance is ensured via beat templates (Sec.~\ref{sec:dataset_generalizability}).

\subsubsection{HP Optimization and Training}
Training was capped at 15 epochs with early stopping (patience = 7) due to rapid convergence. HP (learning rate, batch size, dropout, kernel size $k$, embedding dimension $d$) were optimized with Optuna \cite{Akiba2019} using 5-fold cross-validation on the training set. All single-lead models share the optimal HP from lead II, while ML-Early was optimized independently. For final training, the training dataset was split 80/20 (train/validation) to monitor convergence and limit overfitting. In total, eight single-lead Siamese ResNets and one 8-lead model (ML-Early) were trained.

\subsubsection{Equal Error Rate (EER) as Performance Metric}
The EER, a standard biometric metric, was used as the primary performance measure. It corresponds to the operating point where False Acceptance Rate equals False Rejection Rate, providing a threshold-independent assessment of discriminative performance.

\subsection{Performance on the CPET Dataset: ML-Late Outperforms ML-Early}
\label{sec:perfs}

\begin{figure}[htbp]
\centering

\noindent
\hspace{-0.02\linewidth}
\begin{minipage}[t]{0.715\linewidth}
    \vspace{0pt}
    \includegraphics[width=\linewidth]{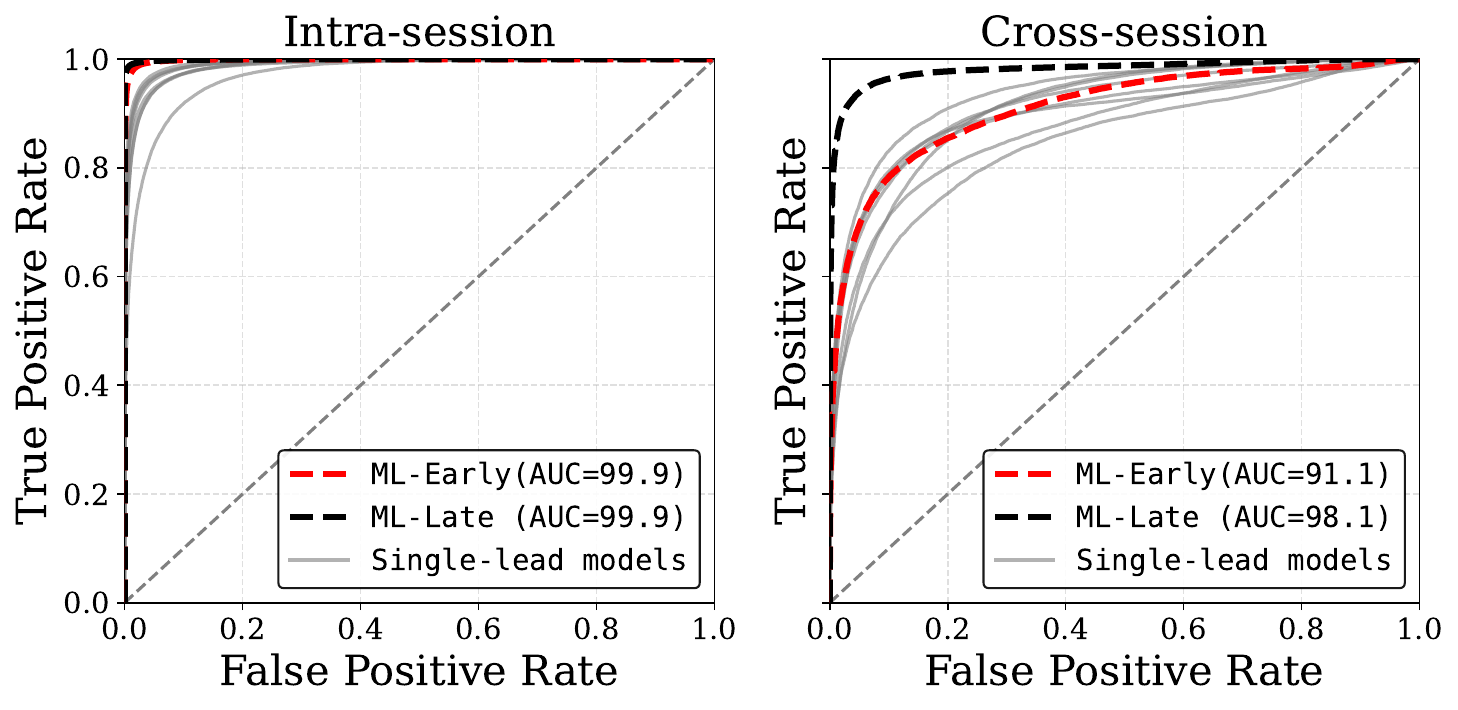}
\end{minipage}
\hspace{0.02\linewidth}
\begin{minipage}[t]{0.26\linewidth}
    \vspace{0pt}
    \setlength{\tabcolsep}{3pt}
    \makebox[\linewidth][r]{%
    \begin{tabular}{l|c|c}
    \cmidrule(lr){2-3}
     & \multicolumn{2}{c}{EER [\%]} \\
    \midrule
    Model & Intra & Cross \\
    \midrule
    ML-Early & 1.7 & 16.4  \\
    ML-Late & \textbf{1.0} & \textbf{5.6}  \\
    Lead II & 5.4 & 13.5  \\
    \bottomrule
    \end{tabular}
    }
\end{minipage}

\caption{
CPET test set performance.
Left: ROC curves (AUC, \%) in intra-session.
Middle: ROC curves in cross-session.
Right: EER (\%) in intra- and cross-session (Lead II is shown as the best single-lead, likely due to its alignment with the heart’s mean electrical axis).
ML-Late outperforms ML-Early, particularly in the cross-session setting.
}

\label{fig:mono_multi_cpet}

\end{figure}

The intra-session and cross-session performance are evaluated on their respective test sets (Sec.~\ref{sec:protocol}). Results are presented on Figure \ref{fig:mono_multi_cpet}.

ML-Late fusion systematically outperforms both single-lead and early multi-lead fusion, achieving EER as low as 1.0\% intra-session. More importantly, it is the only approach showing good performance cross-session with 5.6\% EER. In contrast, ML-Early performs well intra-session (1.7\% EER) but dramatically drops to 16.4\% cross-session—even below the single-lead baseline (Lead II, 13.5\%). For the remainder of the study, only the late fusion model is considered, as it substantially outperforms the other approaches.

\subsection{Cross-Dataset Generalizability: SOTA Performance on Public Benchmarks}
\label{sec:dataset_generalizability}
After evaluating CPET performance, we assess ML-Late generalizability on public datasets against state-of-the-art cross-session results, without dataset-specific fine-tuning. This preserves data in small datasets and ensures a simple, accessible methodology. The same preprocessing is applied across datasets (Sec.~\ref{sec:preproc}).

To ensure full methodological consistency with the ECGXtractor benchmark~\cite{Melzi2023}, we replicated its protocol. Session templates were built by averaging the five beats closest to the mean in Euclidean distance and used as network inputs instead of raw beats. All models were retrained with the same 1:5 genuine-to-impostor ratio. For each dataset, EER was averaged over ten comparison lists: those provided by the benchmark for PTB and CYBHi, and ten generated similarly for Heartprint (which was not included in the benchmark).

Cross-session EERs are reported in Table~\ref{tab:work_comparison}. On PTB, our method achieves 2.1\% EER, matching the ECGXtractor baseline \cite{Melzi2023}. This may be explained by the modest dataset size (113 subjects/templates), which likely promotes stable performance across the ten predefined comparison lists.

\vspace{-0.012 \linewidth}
\begin{table}[b]

\caption{Cross-session EER (\%) on datasets. $^*$ Fine-tuned on target data. Our method matches or outperforms previous results without any fine-tuning.}
\centering

\begin{tabular}{lccc}
\toprule
Work & PTB (12L) & CYBHi (1L) & Heartprint (1L) \\
\midrule
EDITH~\cite{Ibtehaz2022} & 5.7\textsuperscript{*} & - & - \\
\cite{Chee2022} & 10.2 & - & - \\
\cite{Islam2022} & - & - & 53.9 \\
% {\it Proposed} & 3.10 & \textbf{4.43} \\
ECGXtractor benchmark ~\cite{Melzi2023} & \textbf{2.1} & 8.0 (5.4\textsuperscript{*}) & - \\
{\it Proposed} $(\mu$ {\scriptsize $\pm \sigma$}) &
\textbf{2.1}{ \scriptsize$\pm0.3$} &
\textbf{3.9}{ \scriptsize$\pm0.9$} &
\textbf{10.0}{ \scriptsize$\pm1.0$} \\
\bottomrule
\end{tabular}
\label{tab:work_comparison}
\end{table}

The CYBHi and Heartprint datasets are more challenging due to their single finger-to-finger lead. We thus apply the proposed geometry-aware adaptation (Sec.~\ref{sec:Ml}), selecting an ML-Late model based on ${\mathrm{I}, \mathrm{V5}, \mathrm{V6}}$. Lead I captures a horizontal vector in the frontal plane, matching the acquisition geometry, while V5 and V6, though defined in the horizontal plane, form relatively small angles with I and provide complementary, consistent information. Other leads are excluded due to larger angular deviations. For comparison, lead I alone yields EERs of 4.9\% on CYBHi and 10.8\% on Heartprint, versus 4.2\% and 9.8\% using all eight leads. Overall, our method matches ECGXtractor on PTB, significantly outperforms it on CYBHi, and establishes a new benchmark on the less-studied Heartprint dataset, without fine-tuning, achieving state-of-the-art performance.

\subsection{Impact of Exercise}

\begin{figure}[b]
    \centering
    \includegraphics[width=1\linewidth]{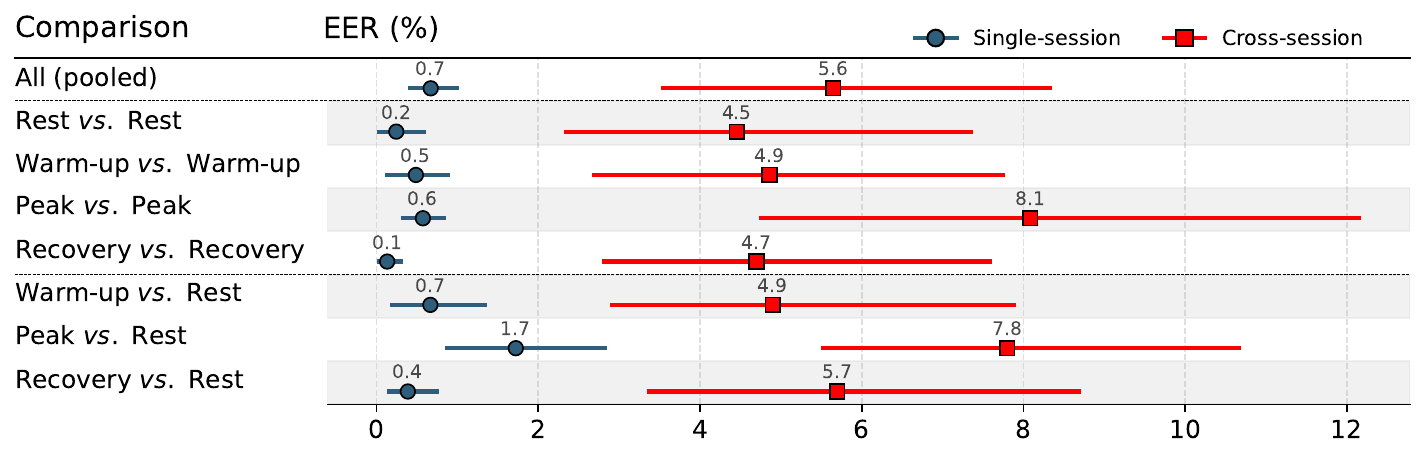}
    \caption{Phase-specific intra- and cross-session EER comparisons, using the 64 multi-session test patients (mean and 95\% CI, intra-session on first test only). For “All”, intra-session EER differs from Sec.~\ref{sec:perfs} due to a different patient subset.}
    \label{fig:EER_exercise_intra}
\end{figure}

One of the main contributions of this work is to assess robustness to exercise-induced ECG physiological variations. Phase-specific evaluation sets were derived from the CPET test subset by selecting beats from rest, warm-up, peak, and recovery phases. Two pairing schemes were considered: (i) same-phase comparisons across tests, and (ii) phase-to-rest comparisons. This was done in both intra- and cross-session scenarios. For all configurations, EER was computed along with 95\% confidence intervals (CI) via 2000-bootstrap resampling.
Given the dependence of bootstrap CIs' on dataset size, evaluation was restricted to the 64-patient multi-session CPET test dataset, using only their first test session for the intra-session configuration.
Results are presented in Figure \ref{fig:EER_exercise_intra}.

\subsubsection{Intra-session}
The pooled intra-session EER (Fig.~\ref{fig:EER_exercise_intra}) confirms excellent overall performance. Phase-specific analysis reveals a moderate drop for \textit{Peak--Rest} (EER 1.7\%, CI [0.8; 2.8]), which remains low, indicating strong robustness to intensity variations. This contrasts with prior studies \cite{Zhou2021,Sung2017,Komeili2016,Cui2021}, which report larger degradations under mixed physiological states, possibly due to smaller datasets. In addition, the \textit{Recovery--Recovery} condition exhibits the lowest error and narrowest CI, an observation of interest given the specific physiological characteristics of this phase, particularly the predominance of vagal tone.

\subsubsection{Cross-session}
Figure \ref{fig:mono_multi_cpet}(right) showed the overall performance decrease to 5.6\%. However, Fig.~\ref{fig:EER_exercise_intra} indicates a sharp increase in variability, with CI widths of at least 5. The largest performance drops are observed for \textit{Peak--Rest} (EER = 7.8\% [5.5; 10.7]) and, interestingly, also for \textit{Peak--Peak} (EER = 8.1\% [4.7; 12.2]).

Three main observations emerge. First, the shift in EER distributions between intra- and cross-session settings—marked by degraded \textit{Peak--Peak} performance and wider CI—indicates a fundamental difference beyond proportional decay, suggesting a non-linear temporal evolution of ECG information. This degradation likely reflects both intrinsic temporal variability and changes in acquisition conditions (e.g., electrode placement, batches, skin preparation, input impedance).

Second, the increased difficulty of \textit{Peak--Peak} in cross-session indicates a loss of discriminative information across recordings. This may be further exacerbated by differences in exercise conditions between sessions, such as variations in workload, pedaling patterns (and associated motion artifacts), as well as changes in baseline physiological state and effort response.

Finally, the similar performance of \textit{Peak--Rest} and \textit{Peak--Peak} shows that ML-Late remains effective even when comparing markedly different physiological states across sessions, a noteworthy contribution. This suggests that, despite variability induced by both physiological changes and acquisition-related factors, the model captures features that retain a degree of invariance across conditions.

\section{Conclusion}
\label{sec:conclusion}

This work proposes ECG-based biometric methods relying on single-lead Siamese ResNet architectures, combined with a late-fusion strategy for multi-lead signals. Trained on an extensive in-house CPET dataset, they outperform state-of-the-art results on PTB and CYBHi (3.94\% EER). It provides the first extensive evaluation of ECG biometrics under exercise conditions, showing strong robustness to both exercise and temporal drift, suggesting the existence of a largely preserved electrocardiographic signature.

However, it is constrained by the scarcity of public datasets combining inter-session variability and physical exertion, and by the limited interpretability of the learned discriminative ECG features, which remains a prerequisite for clinical translation. Future research will focus on transferring representations learned through self-supervised learning-based ECG biometrics (as a pretext task) to prognostic applications, such as mortality and care intensity prediction, alongside improving interpretability.

\begin{credits}
\subsubsection{\ackname} This study was funded by the Excellence Initiative of Aix-Marseille Université (A*Midex, AMX-21-IET-017, “Investissements d’Avenir”).

\end{credits}

\clearpage

% \printbibliography
\bibliographystyle{splncs04}
\bibliography{PhD_bibliography}

\end{document}